\documentclass[10pt]{article} 
\usepackage[preprint]{tmlr}

\usepackage{amsmath,amsfonts,bm}

\def\eqref#1{equation~\ref{#1}}

\def\1{\bm{1}}

\DeclareMathAlphabet{\mathsfit}{\encodingdefault}{\sfdefault}{m}{sl}
\SetMathAlphabet{\mathsfit}{bold}{\encodingdefault}{\sfdefault}{bx}{n}

\usepackage[utf8]{inputenc} 
\usepackage[T1]{fontenc}    
\usepackage{hyperref}       
\usepackage{url}            
\usepackage{booktabs}       
\usepackage{amsfonts}       
\usepackage{nicefrac}       
\usepackage{microtype}      
\usepackage{xcolor}         
\usepackage{comment}
\usepackage{amsmath}

\definecolor{lightred}{RGB}{255, 180, 180}
\definecolor{lightgreen}{RGB}{160, 255, 160}

\title{Soft Guidance Starts to Outperform \\CoT Prompting as LLMs Improve}

\def\month{MM}  
\def\year{YYYY} 
\def\openreview{\url{https://openreview.net/forum?id=XXXX}} 

\author{\name Denys Pushkin \email denys.pushkin@epfl.ch \\
      EPFL, Apple
      \AND
      \name Albert Q. Jiang \thanks{Work done while affiliated with both EPFL and Mistral.} \email albert.jiang@epfl.ch \\
      Mistral AI
      \AND
      \name Aryo Lotfi \thanks{Work done while affiliated with both EPFL and Apple.} \email alotfi@apple.com \\
      Apple 
      \AND
      \name Colin Sandon \email colin.sandon@epfl.ch \\
      EPFL, Apple 
      \AND
      \name Emmanuel Abbé \email e\_abbe@apple.com \\
      EPFL, Apple
      }

\begin{document}

\maketitle

\begin{abstract}
    Chain-of-Thought (CoT) prompting remains the standard baseline for evaluating models' reasoning abilities. Originally, this technique was introduced to elicit step-by-step reasoning from large language models (LLMs), which would otherwise tend to directly output the final answer. However, many modern LLMs produce CoT-style responses \textit{natively} when presented with reasoning tasks, which made us revisit the effectiveness of standard CoT prompting.
    We evaluate several modern mid-sized language models on a math problem-solving task and find that models specialized for reasoning achieve  better performance in a simple zero-shot setting than when using few-shot CoT examples - significantly surpassing officially reported results at no additional cost (e.g., from $\sim$77\% to $\sim$84\% for Mathstral on GSM8K). For the tested general-purpose model, a zero-shot CoT prompt is also sufficient to outperform a few-shot CoT baseline. We attribute this to a `guidance-distraction' tradeoff: standard CoT prompting also demands style adaptation, formatting compliance, and potentially undesired contextualization, which can distract models from the core reasoning task. 
    Our findings suggest that using standard CoT prompting increasingly acts as a source of distraction as models grow stronger.
    
\end{abstract}


\section{Introduction}

Using large language models (LLMs) to solve math problems has become an increasingly active area of research. Chain-of-Thought (CoT) prompting, introduced by \citet{wei2022chain}, is a widely adopted strategy in this domain. In its original formulation, CoT prompting relied on manually crafted examples demonstrating step-by-step reasoning. Building on this, \citet{kojima2022large} proposed zero-shot CoT prompting, which eliminates the need for in-context examples by simply appending the phrase ``Let's think step by step'' to the prompt. Since then, CoT prompting has inspired numerous extensions, while the original CoT prompting became a standard baseline for evaluating reasoning abilities of new models.

The success of CoT prompting was largely attributed to its ability to change the model’s output behavior by encouraging step-by-step rationales rather than direct answers. It mirrors human cognitive process and allows to decompose complex tasks into smaller pieces. However, modern reasoning-specialized LLMs are heavily trained on datasets that already include multi-step reasoning traces, enabling them to produce CoT-style outputs natively.

This raises a natural question: does CoT prompting remain an effective evaluation strategy for models that are already optimized for reasoning tasks? This concern is particularly relevant given that few-shot CoT prompting with human-crafted examples (often sampled from the dataset itself) continues to serve as a standard baseline when evaluating newly released models—such as in the official reports for the models used in this study \citep{mistral2024mathstral, yang2024qwen2}. Re-evaluating the effectiveness of this approach is crucial, as the choice of evaluation baseline plays a critical role in both accurately assessing a model’s true capabilities and fairly estimating the gains from new methods designed to enhance performance.

In this paper, we show that replacing the standard few-shot CoT prompting with zero-shot generation leads to substantial performance gains for reasoning-specialized models. For example, we improve Mathstral’s accuracy from $\leq 74.2\%$ (the best few-shot CoT result in our evaluation with human-crafted examples) or $77.1\%$ (as reported in the original release\footnote{Despite our best effort, we could not reproduce the accuracy stated in the official report of Mathstral.}) to $83.8\%$ using free-form zero-shot generation. This performance further increases to $86.1\%$ when zero-shot CoT prompt is added.

Our main contributions are as follows:

\begin{itemize}
    \item We find that simple zero-shot evaluation significantly outperforms traditional few-shot CoT prompting with randomly selected examples for reasoning-specialized models (e.g., from $\leq 74.1\%$ to $83.8\%$ for Mathstral and from $\leq 83.4\%$ to $88.6\%$ for Qwen). We attribute this to the additional burdens introduced by standard CoT prompting, such as style adaptation and formatting compliance, which can distract models from the core reasoning task. Furthermore, we show that even more advanced CoT prompting strategies that utilize model-generated examples, thus mitigating the previously mentioned challenges, still underperform compared to the zero-shot baseline for Mathstral. These findings suggest that guiding models to produce CoT-style outputs via in-context learning eventually starts to distract more than provide benefits, as the models become more advance in reasoning. 
    
    \item We show that while zero-shot CoT prompting remains effective across the tested models, it provides only marginal gains for reasoning-specialized models. Building on the insights of \citet{kojima2022large}, we argue that the effectiveness of zero-shot CoT prompting is beginning to reach diminishing returns. Drawing an analogy from early developments in AI—where end-to-end training eventually surpassed the use of hand-engineered features—we hypothesize that, as models continue to improve, unconstrained free-form generation will eventually outperform human-crafted prompts such as ``Let's think step by step''.

    \item Based on our findings, we argue that few-shot Chain-of-Thought evaluation can no longer serve as a reliable baseline for reasoning-specialized models, as it may significantly hinder their performance. Persisting with such setups risks systematically underestimating model capabilities and overestimating the gains from alternative methods, which are often benchmarked against this degraded baseline. Instead, we advocate for using unconstrained zero-shot evaluation as a more competitive and faithful measure of reasoning ability in advanced models.



\end{itemize}

\begin{figure}[tb]
\centering
\fbox{%
\begin{minipage}{0.95\linewidth}
\small
\textbf{Problem:} I have 10 liters of orange drink that are two-thirds water and I wish to add it to 15 liters of pineapple drink that is three-fifths water. But as I pour it, I spill one liter of the orange drink. How much water is in the remaining 24 liters?
 \\

\vspace{0.5em}
\textbf{Mathstral-7B Output:} \\
1. First, let's calculate the amount of water in the original 10 liters of orange drink. Since it's two-thirds water, we have:
   Water in orange drink = (2/3) * 10 = 20/3 liters

2. Now, let's calculate the amount of water in the 15 liters of pineapple drink. Since it's three-fifths water, we have:
   Water in pineapple drink = (3/5) * 15 = 9 liters

3. You spill one liter of the orange drink, so the remaining amount of orange drink is:
   Remaining orange drink = 10 - 1 = 9 liters

4. The amount of water in the remaining orange drink is:
   Water in remaining orange drink = (2/3) * 9 = 6 liters

5. Now, let's add the water from the remaining orange drink to the water in the pineapple drink:
   Total water = Water in pineapple drink + Water in remaining orange drink
   Total water = 9 + 6 = 15 liters

6. \colorbox{lightgreen}{Therefore, the final answer is \boxed{15}.}

\vspace{0.5em}
\textbf{Llama-3.1-8B-Instruct Output:} \\
\colorbox{yellow}{16 liters.}  The orange drink is 2/3 water, so it is 2/3 * 9 = 6 liters of water. The pineapple drink is 3/5 water, so it is 3/5 * 15 = 9 liters of water.  \colorbox{lightgreen}{The total amount of water is 6 + 9 = 15 liters}.  Since 1 liter of the orange drink was spilled, the amount of water in the remaining 24 liters is 15 - 1 = 14 liters.  But we know that the total amount of liquid is 24 liters, so the amount of non-water in the remaining 24 liters is 24 - 14 = 10 liters.  The orange drink is 1/3 non-water, so it is 1/3 * 9 = 3 liters of non-water.  The pineapple drink is 2/5 non-water, so it is 2/5 * 15 = 6 liters of non-water.  The total amount of non-water is 3 + 6 = 9 liters.  Since 1 liter of the orange drink was spilled, the amount of non-water in the remaining 24 liters is 9 - 1 = 8 liters.  Since the amount of non-water in the remaining 24 liters is 8 liters, the amount of water in the remaining 24 liters is 24 - 8 = 16 liters.  \colorbox{lightred}{The answer is 16}.  The final answer is 16.  I hope it is correct. 

\end{minipage}%
}
\caption{Example of free-form solutions generated by Mathstral and Llama for the same math problem. Llama initially outputs an incorrect answer (highlighted in yellow), quickly reaches the correct intermediate result (green), but then continues to hallucinate and reaffirms the incorrect final answer (red).}
\label{fig:example_cot}
\end{figure}

\section{Related work}

\citet{wei2022chain} and \citet{kojima2022large} became foundation works that introduced traditional few-show CoT with human-crafted examples and zero-shot CoT approaches. Since then, numerous extensions and refinements to CoT prompting have been proposed.

A substantial body of research has shown that the choice of in-context examples plays a critical role in the effectiveness of CoT prompting \citep{liu2022makes, rubin2022learning, zhou2022least}. As a result, significant effort has been devoted to developing improved example selection methods. These include similarity-based retrieval \citep{rubin2022learning, liu2022makes}, complexity-aware selection \citep{fu2022complexity}, strategies for selecting diverse examples \citep{zhang2022automatic, bai2023transformers, ye2023compositional}, and optimization-based approaches \citep{diao2023active}, among others. This collection of works is complementary to this paper, as we focus on the format of the provided examples for in-context learning (ICL) and use random example selection to clearly isolate the effect of formatting guidance induced by CoT prompting from the confounding effects from advanced example selection techniques.

A more directly relevant line of works investigates alternative formats of in-context examples. Methods such as Auto-CoT \citep{zhang2022automatic}, Active-Prompt \citep{diao2023active}, and AlignedCoT \citep{yang2023alignedcot} propose using model-generated examples, optionally refined through post-processing. These studies show that using model-generated examples can significantly improve the CoT performance. The underlying intuition is that examples produced by the model itself—especially in free-form—better align with its native output style, reducing the burdens of style adaptation and format compliance that often arise with human-crafted examples. In addition, \citet{zhang2022automatic, diao2023active} incorporate zero-shot CoT prompts in example generation stage to further improve the quality of generated examples and performance of the downstream ICL task. 

Our work is concurrent with \citep{cheng2025revisiting} who also study the effectiveness of CoT prompting. Both works agree in the conclusion that zero-shot prompting is more effective that few-shot prompting for modern LLMs on math problems.


\section{Reflections on the effectiveness of Chain-of-Thought prompting}

Chain-of-Thought prompting has become a dominant approach for evaluating and enhancing reasoning in large language models, in particular for math problem solving task.
However, modern instruction-tuned LLMs are trained on datasets rich in step-by-step reasoning examples. These models consistently produce CoT-style outputs even in the absence of explicit prompting. As reasoning models improve, one may need to reexamine whether the structured guidance provided by CoT prompting
continues to serve its original purpose, or it acts like a sources of interference instead.

As an example, Figure~\ref{fig:example_cot} presents solutions generated by the reasoning-specialized Mathstral model and the general-purpose Llama model for the same problem. Both models were given a zero-shot prompt (i.e., only the problem statement), allowing them to generate solutions in their native style.\footnote{The problem was intentionally chosen to expose differences between the models. In some other cases, Llama can also produce high-quality Chain-of-Thought solutions.} Mathstral naturally produces a complete, step-by-step rationale and arrives at the correct answer. In contrast, Llama begins by directly outputting an incorrect final answer and fails to recover. While it briefly reaches the correct intermediate result during reasoning, it ultimately hallucinates further steps and reaffirms its original incorrect answer. This example highlights a key difference between general-purpose and reasoning-specialized models: the former often benefit from explicit guidance via CoT prompting, while the latter are capable of generating coherent multi-step solutions on their own—even in a purely zero-shot setting.

Hence, it is unclear whether reasoning-specialized models still benefit from traditional CoT prompting. These models no longer require external guidance to produce CoT-style rationales ---in fact, such guidance may conflict with the model’s preferred reasoning patterns. Furthermore, they may be hindered by stylistic and formatting constraints imposed by CoT prompt. In this view, standard CoT prompting acts similarly to hand-crafted features in early machine learning pipelines—once useful for compensating for limited model capacity, but eventually outpaced by end-to-end approaches that let the model discover optimal representations independently.


Nevertheless, CoT prompting remains a standard evaluation method for newly released models (e.g., \citet{mistral2024mathstral, yang2024qwen2}), underscoring the need for a careful re-evaluation of their continued relevance. In this work, our goal is to isolate and reexamine the core assumption behind CoT prompting --- that injecting intermediate reasoning structure enhances model performance. To that end, we conduct a controlled study across multiple CoT prompting variants, using random example selection to eliminate confounding effects from retrieval strategies. We compare these strategies against each other and against zero-shot baselines on math problem-solving task using modern mid-sized LLMs.

\section{Experimental setup} \label{sec:exp_setup}

\subsection{Models and dataset}

We evaluate three mid-sized language models with varying levels of specialization in reasoning tasks: \textbf{Mathstral-7B} \citep{mistral2024mathstral}, a model explicitly tuned for multi-step mathematical reasoning; \textbf{Qwen2.5-7B-Instruct} \citep{yang2024qwen2}, a general-purpose model with reported improvements on reasoning benchmarks; and \textbf{Llama-3.1-8B-Instruct} \citep{meta2024llama3}, a general-purpose instruction-following model without specific reasoning optimization.

All models are evaluated on the GSM8K dataset \citep{cobbe2021training}, a widely used benchmark of grade school math word problems. The dataset includes 7.5K training examples and 1.3K test samples. Following standard protocol, we use training examples for in-context demonstration and evaluate performance on the test split.

\subsection{Evaluation methodology}

Our primary goal is to isolate the effect of \textit{reasoning format}—i.e., the structure and style of in-context examples—without confounding effects from advanced example selection techniques. To that end, all few-shot prompting experiments use \textbf{randomly selected examples}. We evaluate a variety of CoT prompting variants that differ in how demonstrations are sourced, formatted, and extracted.

\paragraph{Prompting strategies.}
We study two main classes of prompting methods:
\begin{itemize}
    \item \textbf{Dataset-based examples (DS-CoT)}: Standard few-shot CoT prompting using examples from GSM8K's training set. We consider both versions without and with answer-format instructions, denoted as \textsc{DS-CoT} and \textsc{DS-CoT+Instr}, respectively.
    \item \textbf{Model-generated examples (Self-CoT)}: Few-shot prompting using examples generated by the model itself. We distinguish between \textsc{Self-CoT}, where examples are generated in a free-form zero-shot setting, and \textsc{Self-CoT+0shot}, where examples are generated using a zero-shot CoT prompt (``Let's think step by step'').
\end{itemize}

\paragraph{Answer extraction.}
To decouple reasoning quality from formatting issues, we use two answer extraction methods:
\begin{itemize}
    \item \textbf{Rule-based}: The final answer is extracted using a simple heuristic. Specifically, we expect the model to mimic the dataset’s answer format and search for the last occurrence of the pattern \texttt{\#\#\#\# <number>}.
    \item \textbf{Prompt-based}: The model is explicitly prompted to return only the final answer, stated as a single numerical value.
\end{itemize}
For dataset-based prompts, we evaluate all four combinations of (+Instr) and answer extraction method (+Rule or +Prompt). For model-generated prompts, we use prompt-based answer extraction throughout, since these examples are not explicitly formatted.

\paragraph{Self-CoT prompt construction.}
To construct \textsc{Self-CoT} and \textsc{Self-CoT+0shot} examples, we first generate model solutions for all 7.5K GSM8K training problems—either in free-form or using a zero-shot CoT prompt—and retain only those with correct final answers. In ablation studies (Table~\ref{tab:cleaning_data}), we additionally evaluate a variant that includes a post-processing step to clean the generated demonstrations.

From manual inspection, we identified two primary sources of contamination in model-generated examples: (1) the model continuing a conversation after producing a correct solution (observed for Llama and Qwen), and (2) incomplete reasoning, where the model outputs a correct final answer but fails to logically justify it. To mitigate this, we apply a two-stage autonated cleaning process: first, truncating the output after a complete solution is reached, and second, filtering out examples with potentially incomplete reasoning via a simple automated check.

\paragraph{Implementation details.}
All generations use greedy decoding. Reported results are averaged over 10 runs with different random seeds to ensure robustness to example sampling.

\section{Results}

\begin{table}[tb]
\caption{
Accuracy (in \%) of different CoT prompting strategies and zero-shot baselines on the GSM8K dataset. 
\textbf{DS-CoT} uses dataset examples; \textbf{+Instr} adds answer-format instructions. 
\textbf{Self-CoT} and \textbf{Self-CoT+0shot} use model-generated examples, without and with a zero-shot CoT prompt, respectively. 
\textbf{+Rule} and \textbf{+Prompt} refer to the answer extraction method. 
\textbf{Baselines} include zero-shot free-form prompting and zero-shot CoT.
}
\label{tab:main_results}
    \centering
    \begin{tabular}{c|cc|cc|c|c|c|c}
        \toprule
        & \multicolumn{2}{|c|}{DS-CoT} & \multicolumn{2}{|c|}{DS-CoT + Instr} & \multicolumn{2}{|c|}{Self-CoT} & \multicolumn{2}{|c}{Baslines} \\ 
        \cline{2-9} 
        & +Rule & +Prompt & +Rule & +Prompt & free form & + 0-shot & free form & 0-shot CoT \\ 
        \cline{2-9} 
        & \multicolumn{8}{|c}{Mathstral-7B} \\
        \hline
        0-shot  & -    & -    & - & -    & -    & -    & 83.8 & 86.1 \\
        1-shot  & 60.8 & 72.2 & 72.7 & 72.8 & 81.6 & 84.8 & -    & -    \\
        2-shot  & 71.2 & 72.0 & 73.1 & 73.1 & 82.9 & 84.8 & -    & -    \\
        4-shot  & 72.0 & 72.1 & 73.6 & 73.7 & 83.0 & 85.5 & -    & -    \\
        8-shot  & 73.0 & 73.0 & 74.1 & 74.2 & 83.3 & 85.5 & -    & -    \\
        16-shot & 72.2 & 72.3 & 73.4 & 73.5 & 83.6 & 85.2 & -    & -    \\
        \hline
        & \multicolumn{8}{|c}{Qwen2.5-7B-Instruct} \\
        \hline
        0-shot & -     & -    & - & -    & -    & -    & 88.6 & 90.9 \\
        1-shot & 3.0   & 86.9 & 83.3 & 83.4 & 89.5 & 91.1 & -    &   -  \\
        2-shot & 14.7  & 85.8 & 83.4 & 84.0 & 89.5 & 91.4 & -    & -    \\
        4-shot & 52.8  & 87.6 & 82.5 & 82.8 & 89.6 & 91.6 & -    &  -   \\
        8-shot & 72.9  & 86.4 & 82.5 & 82.7 & 89.4 & 91.5 & -    &  -   \\
        16-shot & 81.1 & 84.5 & 81.1 & 81.3 & 89.6 & 91.6 &  -   &  -   \\
        \hline
        & \multicolumn{8}{|c}{Llama-3.1-8B-Instruct} \\
        \hline
        0-shot  & -    & -    & -    & -    &  -   & -    & 67.3 & 81.4 \\
        1-shot  & 10.6 & 78.0 & 79.0 & 79.6 & 75.8 & 81.1 &  -   & -    \\
        2-shot  & 51.6 & 79.0 & 80.0 & 80.3 & 76.9 & 81.9 &  -   & -    \\
        4-shot  & 72.9 & 80.2 & 79.7 & 80.0 & 77.7 & 81.2 & -    &  -   \\
        8-shot  & 78.4 & 79.8 & 79.0 & 79.3 & 75.8 & 82.0 &  -   & -    \\
        16-shot & 78.8 & 79.3 & 78.9 & 79.1 & 72.8 & 84.4 &  -   &  -   \\
        \bottomrule
    \end{tabular}
\end{table}

\begin{table}[tb]
\caption{
Comparison of accuracy (Acc) and invalid answer rate (Inv Ans, in \%) for different answer extraction methods across CoT prompting variants. 
\textbf{DS-CoT} uses dataset examples; \textbf{+Instr} includes answer-format instructions. 
\textbf{+Rule} applies rule-based answer extraction, while \textbf{+Prompt} uses prompt-based extraction. 
Invalid answers reflect formatting failures—e.g., when the final answer is missing or not expressed as a single number.
}
    \label{tab:unifying_ans_extr}
    \centering
    \begin{tabular}{c|cc|cc|cc|cc}
    \toprule
        & \multicolumn{4}{|c|}{DS-CoT} & \multicolumn{4}{|c}{DS-CoT+Instr} \\
        \cline{2-9}
        & \multicolumn{2}{|c|}{+Rule} & \multicolumn{2}{|c|}{+Prompt} & \multicolumn{2}{|c|}{+Rule} & \multicolumn{2}{|c}{+Prompt} \\ 
        \cline{2-9} 
        & Acc & Inv Ans & Acc & Inv Ans & Acc & Inv Ans & Acc & Inv Ans \\ 
        \cline{2-9} 
        & \multicolumn{8}{|c}{Mathstral-7B-v0.1} \\
        \hline
        1-shot  & 60.8 & 17.3 & 72.2 & 0.6 & 72.7 & 0.8 & 72.8 & 0.6 \\
        2-shot  & 71.2 & 2.0  & 72.0 & 0.6 & 73.1 & 0.7 & 73.1 & 0.5 \\
        4-shot  & 72.0 & 1.0  & 72.1 & 0.5 & 73.6 & 0.6 & 73.7 & 0.5 \\
        8-shot  & 73.0 & 0.7  & 73.0 & 0.6 & 74.1 & 0.6 & 74.2 & 0.6 \\
        16-shot & 72.2 & 0.7  & 72.3 & 0.6 & 73.4 & 0.6 & 73.5 & 0.5 \\
        \hline
        & \multicolumn{8}{|c}{Qwen2.5-7B-Instruct} \\
        \hline
        1-shot & 3.0   & 96.5 & 86.9 & 0.1 & 83.3 & 0.7 & 83.4 & 0.2 \\
        2-shot & 14.7  & 82.4 & 85.8 & 0.1 & 83.4 & 0.8 & 84.0 & 0.2 \\
        4-shot & 52.8  & 39.9 & 87.6 & 0.1 & 82.5 & 0.5 & 82.8 & 0.2 \\
        8-shot & 72.9  & 16.1 & 86.4 & 0.1 & 82.5 & 0.3 & 82.7 & 0.2 \\
        16-shot & 81.1 & 4.6  & 84.5 & 0.1 & 81.1 & 0.2 & 81.3 & 0.2 \\
        \hline
        & \multicolumn{8}{|c}{Llama-3.1-8B-Instruct} \\
        \hline
        1-shot  & 10.6 & 87.3 & 78.0 & 0.1 & 79.0 & 1.7  & 79.6 & 0.0 \\
        2-shot  & 51.6 & 35.7 & 79.0 & 0.1 & 80.0 & 0.8  & 80.3 & 0.0 \\
        4-shot  & 72.9 & 10.0 & 80.2 & 0.1 & 79.7 & 0.7  & 80.0 & 0.0 \\
        8-shot  & 78.4 & 2.4  & 79.8 & 0.1 & 79.0 & 0.5  & 79.3 & 0.0 \\
        16-shot & 78.8 & 1.0  & 79.3 & 0.1 & 78.9 & 0.4  & 79.1 & 0.1 \\
        \bottomrule
    \end{tabular}
\end{table}

We evaluate the prompting variants introduced in Section~\ref{sec:exp_setup}. Results for all evaluated strategies are reported in Table~\ref{tab:main_results}. You can also find the baselines in the last two columns of this table: free-forms means that we only give a problem statement to the model, allowing it to generate its response in completely native format. 0-shot CoT means that we additionally add ``Let's think step by step prefix'' to the model response. Below we list the main finding of our experiments.

\subsection{CoT with human-crafted examples significantly hinders reasoning-specialized models}

We observe that traditional CoT prompting with human-crafted examples—i.e., using demonstrations drawn directly from the training dataset (DS-CoT and DS-CoT+Instr)—consistently underperforms compared to the zero-shot (free form) baseline, regardless of the answer extraction method used (see columns 1–4 in Table~\ref{tab:main_results}). Specifically, we obtain at most $74.1\%$ accuracy with DS-CoT-style prompting for Mathstral, compared to $83.8\%$ for the zero-shot baseline. Similarly, for Qwen, DS-CoT variants achieve no more than $83.4\%$, while zero-shot prompting yields $88.6\%$.

This gap is particularly noteworthy given that few-shot CoT prompting with dataset examples remains the standard evaluation protocol in many official model releases. Our findings show that this approach not only fails to leverage the full reasoning capacity of modern instruction-tuned models—it may actually hinder performance. As models become increasingly capable of generating well-structured rationales natively, rigidly steering them with fixed, human-authored CoT examples can be counterproductive. Persisting with such setups risks systematically underestimating model capabilities and overestimating the gains from alternative methods—since those methods are often benchmarked against a degraded baseline.

\subsection{CoT with model-generated examples does not universally outperform zero-shot baseline}

When comparing \textsc{Self-CoT} (free-form) and \textsc{Self-CoT+0shot} to their respective baselines—zero-shot free-form and zero-shot CoT—we find that these strategies yield only marginal improvements for the Qwen model, and fail to outperform the baseline for Mathstral. We attribute this to Mathstral’s stronger specialization in structured multi-step reasoning. Notably, we distinguish between two related but distinct capabilities: (1) a model’s ability to perform well on reasoning benchmarks, and (2) its proficiency in generating coherent step-by-step rationales. While Qwen achieves higher benchmark accuracy, Mathstral is likely superior in rationale generation, as it was explicitly trained for reasoning tasks.

These findings support our hypothesis that as models become increasingly capable of producing structured solutions natively, imposing additional CoT-style scaffolding may become counterproductive—even when using model-generated demonstrations, which avoid burdens such as style adaptation and formatting compliance. For such models, externally constraining the reasoning format may interfere with their internal problem-solving processes rather than enhance them.

\subsection{\textsc{Self-CoT+0shot} produces the most effective prompt format}

Despite these observations, we note that combining CoT prompting with more advanced example selection strategies remains a promising direction for improving model performance. Our findings do not suggest abandoning CoT prompting entirely, but rather emphasize that its effectiveness depends critically on how examples are selected and formatted. In this context, identifying CoT formats that are minimally intrusive—or, put differently, “least harmful” to the model’s native reasoning behavior—remains an important question.

Among all the prompting variants we evaluated, \textsc{Self-CoT+0shot} (see column 6 in Table~\ref{tab:main_results}) consistently yields the strongest performance across models. It even outperforms the zero-shot CoT baseline for Qwen and Llama, and underperforms it only slightly for Mathstral. These results suggest that model-generated examples, when combined with lightweight CoT-style guidance, offer a more compatible and effective form of prompting. this conclusion aligns with prior work that has successfully adopted similar formats for CoT prompting—often in combination with advanced example selection techniques \citep{zhang2022automatic, diao2023active}.

\subsection{Other observations}

Our results confirm that \textbf{model-generated examples outperform dataset examples for CoT prompting}, provided the model can generate well-structured rationales natively. The only exception is the general-purpose Llama model, where \textsc{Self-CoT} underperforms compared to CoT with dataset examples. This is likely because Llama is not explicitly optimized to produce high-quality reasoning traces. However, when using \textsc{Self-CoT+0shot}, the performance improves and even surpasses dataset-based CoT prompting for Llama. It confirms that using examples in a format non-native to the model hinders performance—an effect that can be attributed to challenges in style adaptation and instruction compliance.

\textbf{DS-CoT+Rule appears to improve with more examples—but this improvement is misleading.} In Table~\ref{tab:main_results}, we observe a large performance gap between \textsc{DS-CoT+Rule} and \textsc{DS-CoT+Prompt}, especially when the number of in-context examples is small. Importantly, the underlying solutions used in both variants are identical; only the answer extraction method differs. This suggests that the gap is not due to better reasoning, but due to more accurate answer extraction. As shown in Table~\ref{tab:unifying_ans_extr}, \textsc{DS-CoT+Rule} requires a large number of examples to consistently produce answers in the expected format. In contrast, \textsc{DS-CoT+Prompt} achieves stable performance with fewer examples, and even exhibits a slight decline with more. This demonstrates that increasing the number of randomly selected examples does not necessarily improve performance—and may misleadingly suggest progress when the real gain is only in formatting compliance.

\textbf{\textsc{Self-CoT+0shot} does not benefit from our post-processing procedure.} We applied a two-step post-processing pipeline to the solutions generated using a zero-shot CoT prompt: (1) truncating the output after the final answer is reached, and (2) filtering out responses that appeared low-quality or logically incomplete, even if the final answer was correct. The effect of this post-processing on downstream ICL performance is shown in Table~\ref{tab:cleaning_data}.

Unlike some prior studies that report performance gains from refining model-generated CoT examples \citep{zhang2022automatic, diao2023active, yang2023alignedcot}, we find that our cleaning procedure does not improve performance for \textsc{Self-CoT+0shot}. The only consistent benefit appears in the \textsc{Self-CoT} (free-form) setting for the Llama model, which is not explicitly tuned for reasoning tasks. While this may partly reflect the simplicity of our post-processing pipeline, it also suggests that such refinement has marginal effect when the model is already capable of producing coherent, well-structured rationales natively.

\begin{table}[tb]
\caption{
Impact of post-processing (cleaning) on CoT prompting performance using model-generated examples across three models on GSM8K. 
“Self-CoT” refers to free-form generation; “Self-CoT+0shot” adds a zero-shot CoT prefix. 
Each cell reports accuracy (\%) and change relative to the unprocessed version. 
}
    \label{tab:cleaning_data}
    \centering
    \begin{tabular}{c|cc|cc|cc|cc|cc|cc}
        \toprule
        & \multicolumn{4}{|c|}{Mathstral} & \multicolumn{4}{|c|}{Qwen} &  \multicolumn{4}{|c}{Llama} \\
        \cline{2-13}
        & \multicolumn{2}{|c|}{\shortstack{Self-CoT\\\phantom{+0shot}}} & \multicolumn{2}{|c}{\shortstack{Self-CoT\\+0shot}}
 & \multicolumn{2}{|c|}{\shortstack{Self-CoT\\\phantom{+0shot}}} & \multicolumn{2}{|c}{\shortstack{Self-CoT\\+0shot}}
 & \multicolumn{2}{|c|}{\shortstack{Self-CoT\\\phantom{+0shot}}} & \multicolumn{2}{|c}{\shortstack{Self-CoT\\+0shot}}
 \\
        \cline{1-13} 
        Cleaning & No & Yes & No & Yes & No & Yes & No & Yes & No & Yes & No & Yes \\
        \hline
        1-shot  & 81.6 & +0.0 & 84.8 & -0.3 & 89.5 & -0.3 & 91.1 & +0.2 & 75.8 & +1.5 & 81.1 & +0.4 \\
        2-shot  & 82.9 & -0.1 & 84.8 & +0.6 & 89.5 & +0.0 & 91.4 & +0.1 & 76.9 & +1.4 & 81.9 & -0.5 \\
        4-shot  & 83.0 & -0.1 & 85.5 & +0.1 & 89.6 & +0.1 & 91.6 & +0.0 & 77.7 & +0.7 & 81.2 & -0.5 \\
        8-shot  & 83.3 & +0.6 & 85.5 & -0.5 & 89.4 & +0.3 & 91.5 & -0.2 & 75.8 & +1.2 & 82.0 & -0.3 \\
        16-shot & 83.6 & +0.1 & 85.2 & +0.0 & 89.6 & +0.2 & 91.6 & +0.0 & 72.8 & +2.2 & 84.4 & -0.4 \\
        \bottomrule
    \end{tabular}
\end{table}

\section{Discussion}

Our results suggest that the effectiveness of Chain-of-Thought (CoT) prompting diminishes as language models become more proficient at generating structured rationales natively. While CoT was once essential for eliciting step-by-step reasoning, modern instruction-tuned models increasingly exhibit these behaviors without explicit guidance.

Although pairing CoT prompting with advanced example selection can still yield performance gains, our findings indicate that such improvements likely stem from the quality of the examples rather than the CoT format itself. In these cases, the benefits of examples relevance outweigh the potential distraction introduced by enforcing a specific reasoning structure.

Zero-shot CoT prompting still offers modest improvements, but these are far smaller than those reported in earlier work \citep{kojima2022large}. We attribute this to the enhanced native reasoning abilities of newer models. Zero-shot CoT prompt may be consider as injecting a human prior into the model’s output, and we anticipate it to face the same fate as hand-crafted features in earlier AI systems — initially helpful, but ultimately outpaced by end-to-end learning (i.e. free form generation in our case).

\paragraph{Limitations and future directions.} Our study focuses on mid-sized models ($\sim$ 7-8B parameters) and uses random selection of in-context examples. While this choice helps isolate the effects of prompting format, more sophisticated example selection methods may interact differently with CoT strategies. Additionally, our experiments are restricted to a single math benchmark (GSM8K); extending this analysis to other reasoning domains, such as symbolic logic or commonsense inference, remains an important direction.

\section{Conclusion}

This study highlights the evolving utility of Chain-of-Thought prompting as language models become more capable reasoners. We demonstrate that traditional few-shot CoT with dataset examples significantly degrades the performance of reasoning-specialized models, and even using model-generated examples could not reach zero-shot baseline for the most advanced reasoning model considered. Using zero-shot CoT prompt remains effective, but produces only marginal gains for models that are already optimized for reasoning.

Our findings point to a guidance–distraction trade-off inherent in CoT prompting. While such prompting guides the model to produce structured, step-by-step rationales, it also introduces challenges such as style adaptation, formatting compliance, and contextual integration, that may distract the model from the core reasoning task. When the guidance offers little or no added value, these distractions begin to outweigh the benefits, ultimately impairing performance.

These results challenge the continued use of few-shot CoT prompting with dataset examples as a default evaluation baseline. This approach can substantially underestimate the abilities of modern LLMs and inflate the perceived benefits of new methods tested against it. We argue that more direct and minimally constrained evaluation strategies —such as zero-shot prompting— offer a more reliable foundation for measuring reasoning abilities. 

\bibliographystyle{plainnat}  
\bibliography{main}

\end{document}